\pdfoutput=1
\documentclass[11pt]{article}
\usepackage{acl}

\usepackage{times}
\usepackage{latexsym}
\usepackage{amssymb}
\usepackage{amsmath}
\usepackage{soul}
\usepackage{hyperref}       
\usepackage{url}            
\usepackage{booktabs}       
\usepackage{amsfonts}       
\usepackage{nicefrac}       
\usepackage{microtype}      
\usepackage{xcolor}         
\usepackage{graphicx}
\usepackage{subfigure}
\usepackage{booktabs}

\usepackage{amsmath}
\usepackage{amssymb}
\usepackage{mathtools}
\usepackage{amsthm}
\usepackage{subfigure}
\usepackage{titlesec}
\usepackage{booktabs} 
\usepackage{makecell}
\usepackage{multirow}
\usepackage{multicol}
\usepackage{diagbox}
\usepackage{caption}
\usepackage{soul}
\usepackage{xcolor}
\usepackage{wrapfig} 
\usepackage{lipsum}  
\usepackage{booktabs}
\usepackage{natbib}
\usepackage{color}
\usepackage{ulem}
\usepackage{microtype}
\usepackage{inconsolata}
\usepackage{subfigure}
\usepackage{graphics}
\usepackage{amsmath,amsfonts,graphicx,amssymb,bm,amsthm,multirow}

\usepackage{booktabs} 
\usepackage{latexsym}
\usepackage{amsmath}
\usepackage{xspace}
\usepackage{amssymb}
\usepackage{graphicx}
\usepackage{multirow}
\usepackage[inline]{enumitem}
\usepackage{makecell}
\usepackage{graphicx}
\usepackage[mathscr]{euscript}
\usepackage{appendix}
\usepackage{algorithm}
\usepackage{algpseudocode}
\usepackage[T1]{fontenc}
\usepackage[utf8]{inputenc}
\usepackage{microtype}
\usepackage{float}
\usepackage{threeparttable}

\usepackage[T1]{fontenc}

\usepackage[utf8]{inputenc}
\usepackage{hyperref}

\usepackage{inconsolata}

\usepackage{graphicx}
\newcommand{\heading}[1]{\vspace*{1mm}\noindent\textbf{#1}}
\title{LINKAGE: Listwise Ranking among Varied-Quality References for Non-Factoid QA Evaluation via LLMs}

\author{Sihui Yang\textsuperscript{\rm 1,2 }\space\space Keping Bi\textsuperscript{\rm 1,2}\thanks{Corresponding
author.}\space\space Wanqing Cui\textsuperscript{\rm 1,2}\space\space Jiafeng Guo\textsuperscript{\rm 1,2}\space\space \textbf{Xueqi Cheng}\textsuperscript{\rm 1,2 }\space\space\\
\textsuperscript{\rm 1}CAS Key Laboratory of Network Data Science and Technology, \\
Institute of Computing Technology, Chinese Academy of Sciences\\
\textsuperscript{\rm 2}University of Chinese Academy of Sciences, Beijing, China\\
\{yangsihui22s, bikeping,cuiwanqing18z, guojiafeng, cxq\}@ict.ac.cn
 \\
}

\begin{document}
\maketitle
\newcommand{\baby}{LINKAGE}
\begin{abstract}

Non-Factoid (NF) Question Answering (QA) is challenging to evaluate due to diverse potential answers and no objective criterion. The commonly used automatic evaluation metrics like ROUGE or BERTScore cannot accurately measure semantic similarities or answers from different perspectives. Recently, Large Language Models (LLMs) have been resorted to for NFQA evaluation due to their compelling performance on various NLP tasks. Common approaches include pointwise scoring of each candidate answer and pairwise comparisons between answers. Inspired by the evolution from pointwise to pairwise to listwise in learning-to-rank methods, we propose a novel listwise NFQA evaluation approach, that utilizes LLMs to rank candidate answers in a list of reference answers sorted by descending quality. Moreover, for NF questions that do not have multi-grade or any golden answers, we leverage LLMs to generate the reference answer list of various quality to facilitate the listwise evaluation. Extensive experimental results on three NFQA datasets, i.e., ANTIQUE, the TREC-DL-NF, and WebGLM show that our method has significantly higher correlations with human annotations compared to automatic scores and common pointwise and pairwise approaches. Our code and dataset can
be found at \url{https://github.com/babyyang525/LINKAGE-Listwise-NFQA-Evaluation}.
\end{abstract}

\section{Introduction}
In recent years, studies on various aspects of Large Language Models (LLMs) have been drawing significant attention, a majority of which are based on the task of factoid question answering (QA) \citep{ARES,searchinthechain,lee2022factuality}.
New evaluation metrics and benchmarks have also been proposed for assessing the factuality of LLMs \cite{wang2023survey,factscore}.
However, much less research has been conducted on non-factoid question answering (NFQA), which usually requires long-form answers to answer open-ended non-factoid questions (NFQ), such as explanations, opinions, or descriptions.
This can be attributed to the inherent difficulty of the NFQA task and the lack of a well-recognized metric to evaluate the generated long-form answers.
Effective evaluation of NFQA is the foundation of developing advanced techniques to enhance the quality of LLMs-generated non-factoid answers.

Evaluating NFQA is challenging since non-factoid questions often involve subjective interpretations and the potential answers can be diverse instead of a definite fact.
Most prior work used automatic evaluation metrics such as measuring word overlaps (e.g., ROUGE \cite{rouge} and BLEU\cite{bleu}) and semantic similarities (e.g., BERTScore \cite{bertscore}) with the ground truth answers. 
To ensure the evaluation reliability, a small amount of manual annotations are also incorporated to compare the NFQA performance. However, both of them have some limitations:
Automatic metrics like ROUGE, BLEU, and BERTScore cannot accurately measure the responses with semantically similar expressions or from a different but reasonable perspective respectively;
Human evaluations, although more accurate in measuring various aspects of the long-form answers, often require annotators to have related knowledge to be reliable and are too expensive to apply on a large scale.\cite{hurdles,webglm}. 
Moreover, even for humans, evaluation of NFQA can still be challenging due to the requirement of domain knowledge as well as subjective interpretations of the questions and judgment criterions.

By ingesting large-scale data from multi-tasks, LLMs, such as the GPT series, have achieved compelling performance on numerous Natural Language Processing (NLP) tasks, and sometimes even outperform humans \cite{llmsurvey}.
Increasing attention has been drawn to leveraging LLMs as surrogates for large-scale evaluation on model-generated responses \cite{factscore,ARES,gptscore}.
Following the routines of human evaluation, approaches that leverage LLMs as judges often adopt the ways of pointwise scoring that grades each candidate answer individually and pairwise comparisons that compare pairs of answers\cite{judgellm}. The pair for comparison can be two candidate answers or a candidate answer and a ground truth answer.
Figure \ref{fig:method} shows a concrete example of these two approaches.

Pointwise grading is hard since the accurate perception of differences between each grade can be difficult. Subtle differences between candidates may not be discerned and reflected in the final score. Pairwise comparison is relatively easier and can be more accurate but it is not scalable to the large number of candidates when the comparisons are between candidates. In contrast, there is no such issue when comparing the pair of a candidate and a ground truth answer. However, it is not feasible when the ground truth is unavailable. Moreover, when only a single ground truth exists, the evaluation may not be accurate to cover various aspects.

Inspired by the evolution of learning to rank in information retrieval, i.e., from pointwise to pairwise to listwise \cite{learning2rank,l2r}, we propose a listwise NFQA evaluation approach that leverages LLMs to conduct ListwIse raNKing AmonG varied-quality referencEs, abbreviated as LINKAGE. Specifically, we use LLMs to assess a candidate answer by its rank in a list of reference answers sorted by quality descendingly. When there are ground truth answers of multiple grades, they can be used as the varied-quality references. When there is only one or no golden answer, we will construct some examples of multi-grade answers and utilize the in-context learning ability of LLMs to generate more reference answers of different quality.
Compared to the pointwise and pairwise approach, listwise ranking can yield more accurate assessment since the LLM judge can take reference answers of various quality into consideration simultaneously. 
When only one reference answer is used, our method degenerates to pairwise comparisons with a ground truth answer. Additionally, given an ordered reference answer list, LLMs only ingest the reference list and candidate answer once, which costs much less than comparing each reference answer with the candidate pairwise and aggregate the score.  

We conduct extensive experiments on three NFQA datasets: ANTIQUE \cite{antique}, the non-factoid portion of TREC DL \cite{trec19,trec20}, and WebGLM \cite{webglm}. ANTIQUE and TREC DL have multi-grade manual annotations on the candidate answers while WebGLM is a non-factoid QA dataset based on Retrieval Augmented Generation (RAG) that provides retrieval passages and a single ground truth answer. Under the settings where there are multiple, single, or none ground truth answers, our method outperforms the automatic similarity scores, as well as pointwise, and pairwise LLM evaluation methods significantly in terms of the correlation with human judgments. By offering more accurate NFQA evaluation, our work can pave the way for future studies on improving NFQA performance, especially promoting LLMs to become more capable of answering complex questions.
\section{Related Work}
\subsection{Non-factoid Question Answering(QA)}
Non-factoid question answering (NFQA) is a complex challenge, characterized by open-ended queries that require complex responses such as descriptions, opinions, or explanations.\citep{Yulianti:2017,nlf6}. 
These responses are usually extensive, often requiring paragraph-level answers.
The most used benchmark in NFQA is the ELI5 dataset \citep{fan2019eli5}, which contains 272,000 questions from the "Explain Like I’m Five" Reddit forum. 
Moreover, multi-document NFQA datasets like WebGLM \citep{webglm}, WikihowQA \citep{wikihowqa} integrate multiple detailed passage-level answers to form long-form answers to NFQ.
ANTIQUE \cite{antique} provides a reliable collection with complete relevance annotations of NFQA.


\subsection{Non-factoid QA Evaluation}
Prior NFQA approaches can be categorized into three categories: 

\begin{figure*}[t]
    \centering
    \includegraphics[width=\linewidth]{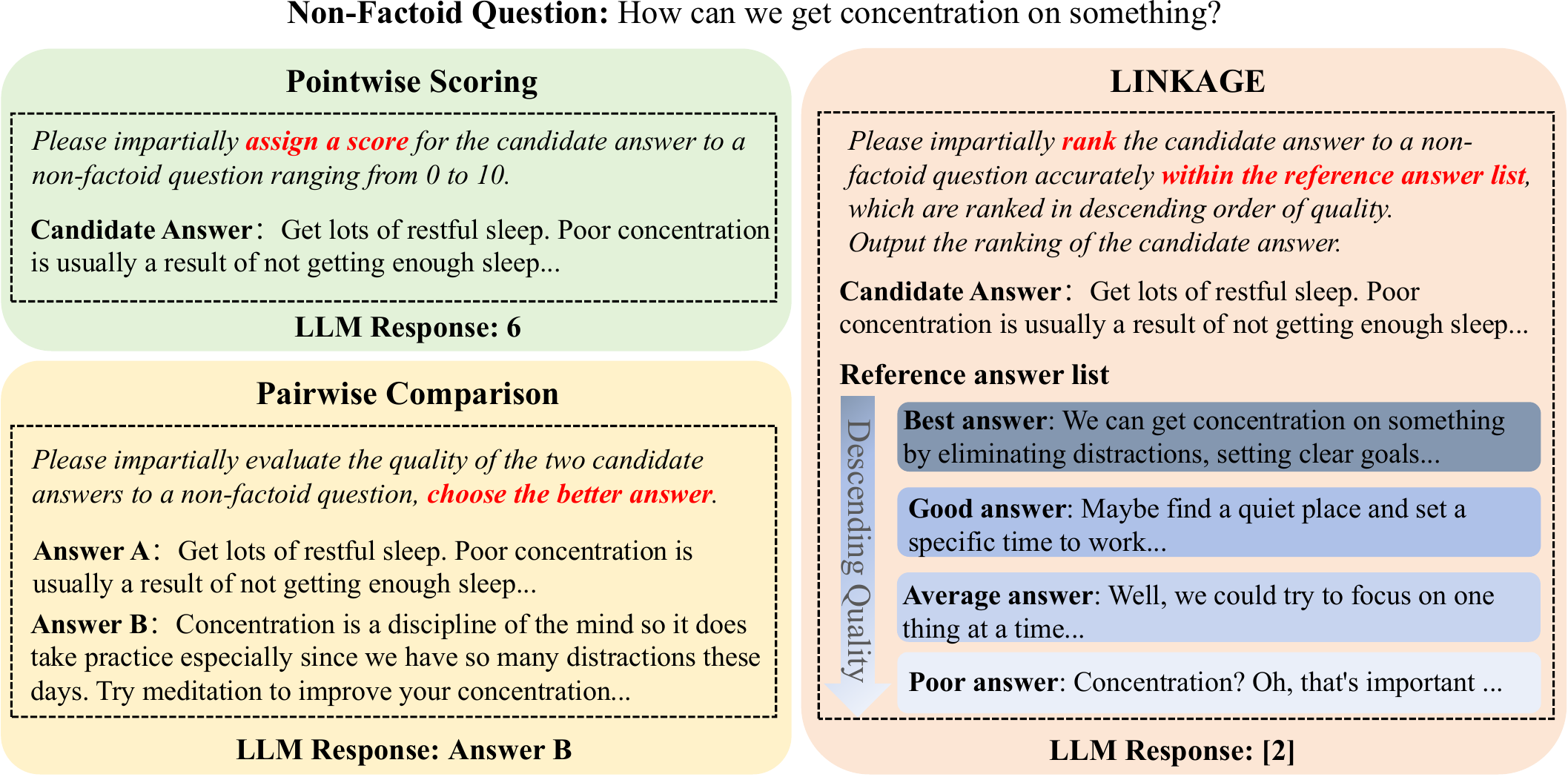}
    \caption{Pointwise scoring evaluation, pairwise comparison evaluation and our \baby\ evaluation approaches.}
    \label{fig:method}
\end{figure*}

\heading{Automatic Evaluation:}
Before the emergence of LLM, the most commonly used evaluation methods were automatic metrics, such as ROUGE \citep{rouge}, BLEU \citep{bleu}, and BERTScore~\citep{bertscore}. These metrics evaluate the quality of a generated answer based on text similarity between the answer and human-written answers.
However, these automatic metrics calculate scores through n-gram similarity, ignoring semantic information. For instance, \citet{hurdles} show that ROUGE is an ineffective metric in long-form question answer tasks.
Another way to implement automatic evaluation is by training a model with human evaluation preferences to conduct automatic assessment, such as QAFactEval\citep{qafacteval} and RankGen\citep{rankgen}. However, these methods struggle to generalize to out-of-domain QA evaluation due to limited human annotations.

\heading{Human Evaluation:}
In NFQA tasks, human annotations are usually considered the golden standard. Hurdles \citep{hurdles}, WebGPT \citep{webgpt}, WikihowQA \citep{wikihowqa} both ask human annotators to choose their preferred answer between the answer generated by the model and the golden answer. Moreover, to compensate for human lack of understanding in certain domains, they can refer to evidence documents during evaluation. However, human evaluation is expensive and therefore difficult to adopt on a large scale.

\heading{LLM Evaluation:}
As LLMs advance, they are gradually replacing costly human annotations. GPTScore \citep{gptscore} uses the generation probability of LLMs to evaluate the model-generated output. 
LLM-Eval\cite{llmeval} uses a unique prompt-based evaluation method for open-domain conversations with LLMs.
PRD \citep{prd} and CHATEVAL \citep{chateval} integrate different LLMs' evaluation results by ranking, discussing, and debating among LLMs. 
The advantage of using LLMs as evaluators lies in their explainability and scalability. However, they also encounter issues such as position bias, verbosity bias, and self-enhancement bias. \citep{judgellm} 
There is a lack of research specifically focused on LLM evaluation for NFQA.

\section{Method}
In this section, we propose a ListwIse raNKing AmonG varied-quality referencEs method (LINKAGE) for evaluating NFQA. 
We formally define the task of NFQA evaluation and introduce some basic evaluation approaches, then introduce the details of our \baby.
\subsection{Preliminary}



\heading{Task Definition:}
\label{task definition}
Given a non-factoid question $q$ and its corresponding n candidate answers $\mathcal{C} = \{c_1, c_2, ..., c_n\}$ to be evaluated, where \( c_i \) represents the \(i\)-th candidate answer. The goal is to score each answer with a scorer $Score(c_i)$. 
The ground truth set of $q$ is $\mathcal{G} = \{ g_1, g_2, ..., g_k \}$, in which $g_i$ represents the  \(i\)-th ground truth.
In this paper, the scorer is LLM and we use a prompt $\mathcal{P}$ to query the LLM to get the scoring results. 

Currently, the commonly used scoring methods based on LLM are pointwise and pairwise approaches\cite{judgellm}.


\heading{Pointwise Evaluation:}
The pointwise evaluation approach assesses an answer $c_i$ only based on its relevance and quality regarding the question $q$.
As shown in Figure \ref{fig:method}, the evaluation process may be conducted with or without using ground truth answers as references.

\begin{equation}
   Score_{\mathrm{point}} (c_i) = f(\mathcal{P}_{\mathrm{point}},q,c_i, \mathcal{R}),
\end{equation}
in which $f(P_{\mathrm{point}}, \cdot)$ represents querying the LLM through prompt $\mathcal{P}_{\mathrm{point}}$. 
$\mathcal{R}= [r_1, r_2, \ldots, r_m]$ is a reference answer list sorted by quality in descending order, which can be $\mathcal{G}$, a subset of $\mathcal{G}$, or $\emptyset$.

Pointwise grading is easy to conduct but difficult to accurately perceive grade differences. The subtle differences among candidates may not be distinguished and reflected in the final score.


\heading{Pairwise Evaluation:}
As shown in Figure \ref{fig:method}, the pairwise evaluation approach performs a pairwise comparison between answers. The pairs can be two candidate answers, 
\begin{equation}
    Score_{\mathrm{pair}} (c_i) = \sum_{c_j \in \mathcal{C} \setminus \{c_i\}} f(\mathcal{P}_{\mathrm{pair}},q, c_i, c_j).
\end{equation}
However, the number of comparisons between candidate answer pairs grows exponentially with the number of candidate answers, and thus cannot be scaled to a large number of candidates.
The pair can also be a candidate answer and a reference answer,
\begin{equation}
    Score_{\mathrm{pair}} (c_i) = \sum_{r_j \in \mathcal{R}} w_{l_j} * f(\mathcal{P}_{\mathrm{pair}},q, c_i, r_j),
\end{equation}
\begin{equation}\label{list}
    f(\mathcal{P}_{\mathrm{pair}},q, c_i, r_j) = \left \{
    \begin{aligned}
    1 & \text{, if }c_i\text{ is better} \\
    -1 & \text{, if }r_j\text{ is better} \\
    0 & \text{, otherwise} \\
    \end{aligned}
    \right..
\end{equation}
$\mathcal{R}$ can be $\mathcal{G}$ or a subset of $\mathcal{G}$. 
$w_{l_j}$ is the weight corresponding to certain grade $l_j$ of answer $r_j$. In this way, the pairwise approach scores a candidate answer by comparing it with each answer in the reference answer list.



Pairwise comparison is relatively easier and can be more accurate, but when there is only a single ground truth, evaluation becomes less accurate because it is difficult for a single ground truth to cover various aspects of NFQA

\subsection{Listwise Ranking Evaluation (\baby)}
    

Figure~\ref{fig:method} shows how our \baby\ works. Specifically, given a reference answer list sorted by descending quality and the answer to be evaluated, the scorer judges its quality by deciding where it should be ranked among the reference answer list,
\begin{equation}
  Score_{\mathrm{pair}} (c_i) = f(\mathcal{P}_{\mathrm{list}},q, c_i, \mathcal{R}).
\end{equation}
The higher the ranking, the better the quality.

Please note the difference between our method and the pointwise approach with references. Although both methods ask LLMs to directly output a numerical value, in the pointwise approach, references are used to provide a criterion for scoring, and the assignment only focuses on the quality of $c_i$ itself rather than comparisons. The listwise ranking approach relies on comparing it with all reference answers to determine where the answer should be ranked.

\subsection{Reference List Construction}
Reference answer list $\mathcal{R}$ in \baby\ is composed of multiple answers ordered in descending quality. Compared to providing LLMs with only one ground truth, more references with different styles and quality enable the LLM evaluators to learn implicit evaluation guidelines from $\mathcal{R}$.
The collection method of $\mathcal{R}$ depends on the composition of the ground truth set of the dataset, and we discuss it in three situations:

\subsubsection{Multi-grade Ground Truth}
When multiple grades of ground truth answers are available, references can be sampled directly from these answers. For instance, ANTIQUE and TREC DL contain multiple answers annotated with four relevant labels.

To reduce bias and ensure the reliability of evaluation results, we randomize the sampling process multiple times. 
Additionally, the length and the distribution of $\mathcal{R}$ also impact the results.
We discuss this in detail in Section~\ref{composition of R}.

\subsubsection{Single-grade Ground Truth}
\label{single-grade answer}
\label{1gt}
Some NFQA datasets, such as WebGLM, only contain a single grade of ground truth. 
For this scenario, we prompt LLMs to generate answers of varying quality to serve as references.
Specifically, we first prompt LLMs to answer the question based on the original golden answer, thus obtaining a new high-quality golden answer. The prompt is in Figure \ref{fig:generate r4} (Appendix \ref{a.2}). This step ensures that both the golden reference and other reference answers are generated by LLMs, avoiding the introduction of style bias between human and machine writing. We then use the prompt in Figure~\ref{fig:generate other R} (Appendix \ref{a.2}) to obtain other lower-quality reference answers.
To ensure the diversity of references, we use three LLMs to generate separate lists of reference answers. Then we randomly sample reference answers from three lists to form $\mathcal{R}$ for each grade. 


\subsubsection{Absence of Ground Truth}
In real-world scenarios, non-factoid questions may not have reference answers.
To tackle the problem of ground truth missing, considering the powerful capabilities of LLMs like GPT-4~\cite{gpt4}, we get a quality-assured answer from GPT-4 directly.
The ways of generating reference answers of other quality are the same as described in Section \ref{1gt}.


\section{Experimental Settings}
\subsection{Datasets}
We evaluate the effectiveness of baseline methods and our proposed \baby\ using the following three datasets. 

\textbf{ANTIQUE~\cite{antique}} dataset contains 2,626 open-domain non-factoid questions asked by real users in a community question answering service, i.e., Yahoo! Answers. 
Similar to TREC-DL, all passages are graded into four levels (3: reasonable and convincing, 2: not sufficiently convincing, 1: unreasonable, 0: make no sense). 
We merge the 200 questions from the test set and the 300 questions randomly sampled from the training set, yielding a total of 500 queries as our experiment dataset.

\begin{table}[t]
\setlength{\abovecaptionskip}{0pt}
    \caption{Statistics of ANTIQUE and TREC-DL-NF we use in experiments.}
\renewcommand{\arraystretch}{1}
    \setlength\tabcolsep{2pt}
    \centering
    \label{statistic}
    \begin{tabular}{lcccccc}
    \toprule
        Number statistics & ANTIQUE & TREC-DL-NF \\
        \midrule
        {\#Question}  & 500 & 55 \\
        {\#Avg doc labeled 3}  & 5.8 & 9.6 \\
        {\#Avg doc labeled 2}  & 4.5 & 18.1 \\
        {\#Avg doc labeled 1}  & 6.5 & 24.9 \\
        {\#Avg doc labeled 0}  & 3.6 & 48.0 \\
        {\#Avg total documents}  & 20.4 & 100.7 \\
    \bottomrule
    \end{tabular}
\end{table}

\textbf{TREC-DL-NF~\cite{trec19,trec20}} 
In our experiments, we use TREC-DL 2019, 2020 datasets, which comprise 43 and 54 MS MARCO queries respectively. Each question has multiple passages labeled with four levels of relevance (3: perfectly relevant, 2: highly relevant, 1: related, 0: irrelevant).
Not all questions are NF questions, so we filter factoid questions with a non-factoid question category classifier \cite{taxonomy}. 
This leaves us a total of 55 non-factoid questions, denoted as TREC-DL-NF.

The statistics of ANQIQUE and TREC-DL-NF can be found in Table~\ref{statistic}.

\textbf{WebGLM~\cite{webglm}} is a high-quality quoted long-formed retrieval-augmented QA dataset. 
Each question is accompanied by 5 top-ranked documents retrieved by a vanilla Contriever \cite{contriever}. Question and corresponding candidate references are fed together to OpenAI text-davinci-003~\cite{gpt3} to generate long-formed answers by 1-shot in-context learning.
To obtain candidate answers of different styles and quality, we use gpt-3.5-turbo-16k ~\cite{chatgpt} to generate two answers with 5 relevant and 3 relevant plus 2 irrelevant documents respectively. The third answer is generated by Mistral-7B-Instruct-v0.2~\cite{jiang2023mistral} with 5 relevant documents.
We sample 50 cases and manually label three candidate answers with three levels (3,2,1). Details about manual annotation are in the Appendix \ref{appendix:human}.

\subsection{Methods for Comparison}
\label{methods for comparison}
We compare the following NFQA evaluation baselines and our \baby\ under different situations.
\subsubsection{Baselines} 
\heading{Automatic Metrics:} 

\textbf{ROUGE}\cite{rouge}, \textbf{BERTScore}\cite{bertscore}, \textbf{BLEU}\cite{bleu} are all reference-based metrics based on text similarity. ROUGE and BLEU focus on exact n-gram matching, while BERTScore evaluates the semantic similarity of embeddings.
    
\heading{LLM Evaluation Baselines:}

\begin{itemize}[leftmargin=*,itemsep=0pt,topsep=0pt,parsep=0pt]

    \item \textbf{Pointwise$^{R=\emptyset}$}: This method asks LLMs to directly assign a quality score from 1 to 10 to the candidate answer without any reference answers.
    \item \textbf{Pointwise$^{R \neq \emptyset}$}: Based on the basic pointwise method, this method also provides a list of reference answers sorted in descending order of quality for LLMs to refer to when scoring.
    \item \textbf{Pairwise}: This method scores a candidate answer based on comparing it with each answer in the reference list.
    To eliminate position bias, i.e., the LLM judge might favor the forward-positioned one when comparing two answers, we randomly permute the positions of the candidate answer and ground truth answer during evaluation.
\end{itemize}
\begin{table*}[ht]
\setlength\tabcolsep{10pt}
\renewcommand{\arraystretch}{1}
\setlength{\abovecaptionskip}{0pt}
\caption{The performance of different methods on ANTIQUE and TREC-DL-NF. K, S, and P represent Kendall’s tau, Pearson's r, and Spearman's rho coefficient respectively.
The best results of each evaluator model are in bold.
}
\label{main_exp}
\resizebox{\textwidth}{!}{
\large
\begin{threeparttable}
  \begin{tabular}{clcccccc}
    \toprule
    & \multirow{2}{*}{Method} & \multicolumn{3}{c}{ANTIQUE} & \multicolumn{3}{c}{TREC-DL-NF} \\
    \cmidrule(lr){3-5} \cmidrule(lr){6-8}
    & & K & S & P & K & S & P \\
    \midrule
    \multirow{5}{*}{\makecell[c]{\text{Automatic} \\ \text{Metrics}}} & ROUGE-1 & 0.2088 & 0.2563 & 0.2878 & 0.2442 & 0.3060 & 0.3412 \\
    & ROUGE-2 & 0.1807 & 0.2089 & 0.2281 & 0.2064 & 0.2441 & 0.2808 \\
    & ROUGE-L & 0.2012 & 0.2463 & 0.2708 & 0.2171 & 0.2721 & 0.3178 \\
    
    & BERTScore & 0.1562 & 0.1938 & 0.1950 & 0.2258 & 0.2824 & 0.2842 \\
    
    & BLEU & 0.1808 & 0.2153 & 0.2063 & 0.2106 & 0.2650 & 0.2208 \\
    \midrule
    \multirow{3}{*}{\makecell[c]{\text{LLM Evaluation} \\ \text{on Mistral}}} 
    & Pointwise$^{R=\emptyset}$ & 0.2202 & 0.2499 & 0.2519 & 0.2366 & 0.2773 & 0.2677 \\
    
    & Pointwise$^{R\neq \emptyset}$ & 0.2229 & 0.2516 & 0.2547 & 0.3138 & 0.3382 & 0.3302 \\
    
    & Pairwise & 0.1827 & 0.2134 & 0.2132 & 0.2501 & 0.2967 & 0.2939 \\
    \midrule
    \multirow{2}{*}{\makecell[c]{\text{\baby} \\ \text{on Mistral}}} 
    & \baby$^{0\_shot}$ & 0.3585  & 0.3790  & 0.3893 & 0.3287  & 0.3539  & 0.3401 \\
    
    & \baby$^{few\_shot}$ & \textbf{0.3742}  & \textbf{0.4200}  & \textbf{0.4373} & \textbf{0.4312}  & \textbf{0.4725}  & \textbf{0.4958} \\
    \midrule
    \multirow{3}{*}{\makecell[c]{\text{LLM Evaluation} \\ \text{on ChatGPT}}} 
    & Pointwise$^{R=\emptyset}$ & 0.2777 & 0.3118 & 0.3244 & 0.3176 & 0.3640 & 0.3660 \\
    
    & Pointwise$^{R\neq \emptyset}$ & 0.2752 & 0.3112 & 0.3224 & 0.3746 & 0.4288 & 0.4449 \\
    
    & Pairwise & 0.2979 & 0.3494 & 0.3756 & 0.3204 & 0.3692 & 0.3749 \\
    \midrule
    \multirow{1}{*}{\makecell[c]{\text{\baby} \\ \text{on ChatGPT}}} 
    & \baby$^{0\_shot}$ & 0.3070  & 0.3404  & 0.3514 & 0.3923  & 0.4315 & 0.4376 \\
    & \baby$^{few\_shot}$ & \textbf{0.3096}  & \textbf{0.3543}  & \textbf{0.3688} & \textbf{0.3993}  & \textbf{0.4325}  & \textbf{0.4481} \\
    
    
    \bottomrule
  \end{tabular}
\end{threeparttable}
}
\end{table*}

    
    
    
    
    
\begin{table}[ht]
\setlength\tabcolsep{2pt} 
\renewcommand{\arraystretch}{1.2} 
\caption{Results for the situation of single-grade ground truth. The best results of each model are in bold.}
\label{1GT}
\resizebox{\columnwidth}{!}{%
\begin{threeparttable}
  \begin{tabular}{lccccc}
    \toprule
     \multirow{2}{*}{Model} & \multirow{2}{*}{Method} & \multicolumn{2}{c}{ANTIQUE} & \multicolumn{2}{c}{TREC-DL-NF} \\
    \cmidrule(lr){3-4} \cmidrule(lr){5-6}
    & & K & S & K & S \\
    \midrule
    
    \multirow{5}{*}{Mistral} 
    & Pointwise$^{R=\emptyset}_{1GT}$ & 22.02 & 24.99 & 23.66 & 27.73\\
    & Pointwise$^{R\neq \emptyset}_{1GT}$ & 25.26 & 28.31 & 33.28 & 38.25 \\
    & Pairwise$_{1GT}$ & 20.89 & 23.41 & 30.43 & 36.62 \\
    \cmidrule(lr){2-6}
    & \baby$^{0\_shot}_{1GT}$ & 32.92 & 35.80 & 36.60 & 39.93 \\
    & \baby$^{few\_shot}_{1GT}$ & \textbf{42.89} & \textbf{47.06} & \textbf{42.13} & \textbf{46.18} \\
    \midrule
    \multirow{4}{*}{ChatGPT}
    & Pointwise$^{R=\emptyset}_{1GT}$ & 27.77 & 31.18 & 31.76 & 36.40\\
    
    & Pointwise$^{R\neq \emptyset}_{1GT}$ & 27.91 & 30.71 & 39.75 & 44.66 \\
    
    & Pairwise$_{1GT}$ & 29.88 & 32.32 & 30.28 & 34.14 \\
    \cmidrule(lr){2-6}
    & \baby$^{few\_shot}_{1GT}$ & \textbf{32.93} & \textbf{33.54} & \textbf{44.83} & \textbf{48.51} \\
    \bottomrule
  \end{tabular}
\end{threeparttable}
}
\end{table}

\begin{table}[ht]
\setlength\tabcolsep{2pt} 
\renewcommand{\arraystretch}{1.2} 
\setlength{\abovecaptionskip}{0pt}
\caption{Results for the situation of absence of ground truth. The best results of each model are in bold.}
\label{0GT}
\resizebox{\columnwidth}{!}{%
\large
\begin{threeparttable}
  \begin{tabular}{lccccc}
    \toprule
    \multirow{2}{*}{Model} & \multirow{2}{*}{Method} & \multicolumn{2}{c}{ANTIQUE} & \multicolumn{2}{c}{TREC-DL} \\
    \cmidrule(lr){3-4} \cmidrule(lr){5-6}
    & & K & S & K & S \\
    \midrule
    \multirow{3}{*}{Mistral} 
    & Pointwise$^{R=\emptyset}_{0GT}$ & 22.02 & 24.99 & 23.66 & 27.73\\
    \cmidrule(lr){2-6}
    & \baby$^{0\_shot}_{0GT}$ & 30.05 & 32.87 & 34.28 & 37.65\\
    & \baby$^{few\_shot}_{0GT}$ & \textbf{39.51} & \textbf{43.48} & \textbf{42.35} & \textbf{46.39}\\
    \midrule
    \multirow{2}{*}{ChatGPT} 
    & Pointwise$^{R=\emptyset}_{0GT}$ & 27.77 & 31.18 & 31.76 & 36.40\\
    \cmidrule(lr){2-6}
    & \baby$^{few\_shot}_{0GT}$ & \textbf{36.57} & \textbf{40.43} & \textbf{43.77} & \textbf{46.96}\\
    \bottomrule
  \end{tabular}
\end{threeparttable}
}
\end{table}

\subsubsection{\baby} 
\textbf{\baby:} To ensure that $\mathcal{R}$ uniformly contains answers of varying quality, we randomly select the same number of reference answers from the answer set of each level to create the reference answer list.
For TREC-DL-NF, the grades of answers in the reference list are $\mathcal{L}=(3,2,1,0)$. For ANTIQUE, $\mathcal{L}=(3,3,2,2,1,1,0,0)$, which are intuitively reasonable and balanced settings.

\textbf{\baby-1GT:} We also test the case where there is only one ground truth. For questions with multi-grade answers, we randomly sample one answer from the highest-grade ground truth set as the only ground truth to simulate this situation.

\textbf{\baby-0GT:} In this case, we do not use any labeled ground truth to simulate the situation where no ground truth is available.

\subsection{Evaluation Metrics}
To evaluate the effectiveness of NFQA evaluation, we use \textbf{Kendall's tau}, \textbf{Pearson's r} and \textbf{Spearman's rho coefficient} to calculate the extent of consistency between the resulting sorted sequences and the manually labeled sequences. Spearman's rho coefficient is chosen as our primary metric due to its balance between robustness and sensitivity to monotonic relationships.
\subsection{Implementation Details}
The evaluation experiments are based on two representative LLMs: \begin{enumerate*}[label=(\roman*)]
    \item The open-source model Mistral (Mistral-7B-Instruct-v0.2~\footnote{\url{https://huggingface.co/mistralai/Mistral-7B-Instruct-v0.2}})~\cite{jiang2023mistral}. 
    \item The close-source model ChatGPT (gpt-3.5-turbo-16k)~\cite{chatgpt}, for which results are obtained through API.
\end{enumerate*}
The temperature for all experiments is set to 0.8.

When only one or no ground truth exists, we use gpt-4-1106-preview \cite{gpt4} to generate the golden answer.
For generating other references with descending quality, we use three different LLMs in 3-shot setting: \begin{enumerate*}[label=(\roman*)]
    \item Mistral-7B-Instruct-v0.2,
    \item gpt-3.5-turbo-16k,
    \item Meta-Llama-3-8B-Instruct~\footnote{\url{https://huggingface.co/meta-llama/Meta-Llama-3-8B-Instruct}}~\cite{llama3}.
\end{enumerate*}
All our experiments are done on a single Tesla A100 80G GPU.

\section{Experimental Results}\label{section:results}

\begin{table}[t]
    \caption{Results on WEBGLM based on Mistral. RL, BS, and B represent ROUGE-L, BERTScore, and BLUE, respectively. Acc(b) means the accuracy of finding the best answer. Acc(b+w) means the accuracy of finding both the best and the worst answers.}
    \label{webglm}
    \renewcommand{\arraystretch}{1.3}
    \centering
    \large  
    \setlength\tabcolsep{3pt}  
    \resizebox{\columnwidth}{!}{%
    \begin{tabular}{lcccccc}
        \toprule
        & RL & BS & B  & Point$^{R\neq \emptyset}$ & Pair & \baby \\
        \midrule
        \text{Acc(b)} & 0.42 & 0.48 & 0.50 & 0.46 & 0.54 & \textbf{0.76} \\
        \midrule
        \text{Acc(b+w)} & 0.32 & 0.38 & 0.44 & 0.22 & 0.32 & 0.34 \\
        \bottomrule
    \end{tabular}
    }
\end{table}

\begin{table}[t]
\setlength\tabcolsep{5pt} 
\renewcommand{\arraystretch}{1.2} 
\setlength{\abovecaptionskip}{0pt}
\caption{Different composition of $\mathcal{R}$ on ANTIQUE and TREC-DL-NF and  using Mistral.
The settings we use in \baby\ are in bold.
}
\label{composition effect}
\resizebox{\columnwidth}{!}{ 
\large 
\begin{threeparttable}
  \begin{tabular}{>{\centering\arraybackslash}p{1.5cm} >{\centering\arraybackslash}p{0.5cm} >{\centering\arraybackslash}p{1.5cm} cc cc}
    \toprule
    \multirow{2}{*}{Dataset} & \multirow{2}{*}{|$\mathcal{R}$|} & \multirow{2}{*}{$\mathcal{R}$} & \multicolumn{2}{c}{0-shot} & \multicolumn{2}{c}{3-shot} \\
    \cmidrule(lr){4-5} \cmidrule(lr){6-7} 
    & & & K & S & K & S\\
    \midrule
    \multirow{4}{*}{\rotatebox[origin=c]{90}{ANTIQUE}} 
    & 4 & 3210 & 24.68 & 25.95 & 26.79 & 29.65\\
    \cmidrule(lr){2-7}
    & \multirow{3}{*}{8} & \textbf{33321000} & \textbf{29.88} & \textbf{32.62} & \textbf{31.54} &\textbf{ 34.81} \\
    & & 33221100 & 21.40 & 22.93 & 25.01 & 27.18\\
    & & 32221110 & 25.77 & 29.16 & 28.10 & 29.65\\
    \midrule
    \multirow{4}{*}{\rotatebox[origin=c]{90}{TREC-DL-NF}} 
    & 4 & 3210  & 25.79 & 27.52 & 31.78 & 34.83\\
    \cmidrule(lr){2-7}
    & \multirow{3}{*}{8} & 33321000 & 24.03 & 25.81 & 30.90 & 34.24\\
    & & \textbf{33221100} & \textbf{25.85} & \textbf{27.61} & \textbf{32.70} & \textbf{36.30} \\
    & & 32221110 & 24.94 & 27.01 & 30.97 & 34.39\\
    \bottomrule
  \end{tabular}
\end{threeparttable}
}
\end{table}

\subsection{Overall Results}
The results on the multi-grade ground truth situation, single-grade ground truth situation, and absence of ground truth situation are shown in Table~\ref{main_exp}, Table~\ref{1GT}, and Table~\ref{0GT} respectively. The results on WebGLM are shown in Table~\ref{webglm}. It can be seen that our method always shows better consistency with human evaluation.

Additionally, we have the following observations: 

\textbf{LLM-based methods perform generally better than automatic metrics.}
This indicates that automatic metrics have limitations in NFQA evaluation, therefore should be used with caution in future research. Among LLM-based methods, our proposed \baby\ outperforms all other baselines by a significantly large margin leveraging both Mistral and ChatGPT. This confirms the superiority of listwise approach over the pointwise and pairwise approaches on NFQA evaluation. 

\textbf{Few-shot in-context Learning can enhance the performance of \baby.}
Comparing with results under few-shot and zero-shot, providing LLMs with a few examples can help demonstrate the evaluation task more clearly.
Compared to Mistral, the enhancement of few-shot ICL on ChatGPT is less. We think that it is because ChatGPT has a much better understanding of instructions so the few-shot example does not help it much.

\begin{table}[t]
  \centering
    \caption{The performance of \baby\ under different few shot setting on TREC-DL-NF using Mistral.}
    \begin{tabular}{cccc}
    \toprule
        n-shot  & Kendal& Spearman  & Pearson\\
    \midrule
    $n$=0   & 0.3287  & 0.3539  & 0.3401  \\
    $n$=1   & 0.4246  & 0.4656  & 0.4724  \\
    $n$=3   & \textbf{0.4312}  &  \textbf{0.4752}  & \textbf{0.4958}  \\
    $n$=5   & 0.4339  & 0.4725  & 0.4958  \\
    \bottomrule
    \end{tabular}%
  \label{tab:fewshot_trec}%
\end{table}

\begin{table}[t]
  \centering
    \caption{The performance of \baby\ under different few shot setting on ANTIQUE using Mistral.}
    \begin{tabular}{cccc}
    \toprule
        n-shot  & Kendal& Spearman  & Pearson\\
    \midrule
    $n$=0   & 0.2900  & 0.3101  & 0.3172  \\
    $n$=1   & \textbf{0.3850}  & \textbf{0.4183}  & \textbf{0.4256 } \\
    $n$=3   & 0.3696 &  0.4012  & 0.4122  \\
    $n$=5   & 0.3654  & 0.3934  & 0.4041  \\
    \bottomrule
    \end{tabular}%
  \label{tab:fewshot_antique}%
\end{table}

However, the number of samples cannot be too large. We conduct several sets of few-shot experiments on TREC-DL-NF and ANTIQUE using Mistral. As the results are shown in Table \ref{tab:fewshot_trec} and Table \ref{tab:fewshot_antique}. When the number exceeds a certain value, the performance will deteriorate. This is because the shot number increasing leads to a significant increase in the input length, which will make the LLMs difficult to understand.



\begin{table*}[t]
\setlength\tabcolsep{2pt} 
\renewcommand{\arraystretch}{1.2} 
\setlength{\abovecaptionskip}{0pt}
\caption{Average Spearman coefficient and standard deviation of randomly selecting $\mathcal{R}$ in three dependent experiments on TREC-DL-NF using Mistral and ChatGPT.}
\label{tab:random trec}
\centering
\begin{threeparttable}
  \begin{tabular}{lcccccc}
    \toprule
    Model & Method & Spearman 1 & Spearman 2 & Spearman 3 & Average & Std \\
    \midrule
    \multirow{3}{*}{Mistral} 
    & Pointwise$^{R\neq \emptyset}$ & 0.3382 & 0.3463 & 0.3567 & 0.3471 & 0.0093\\
    & Pairwise &0.2967 & 0.2783 & 0.2912 & 0.2887 & 0.0094\\
    & \baby$^{few\_shot}$ & 0.4725 & 0.4579 & 0.4520 & 0.4608 & 0.0105\\
    \midrule
    \multirow{3}{*}{ChatGPT} 
    & Pointwise$^{R\neq \emptyset}$ & 0.2777 & 0.4288 & 0.4526 & 0.3864 & 0.0948\\
    & Pairwise &0.3692 & 0.3687 & 0.3544 & 0.3641 & 0.0083\\
    & \baby$^{few\_shot}$ & 0.4094 & 0.3854 & 0.4325 & 0.4091 & 0.0235\\
    \bottomrule
  \end{tabular}
\end{threeparttable}
\end{table*}

\textbf{Reference answer list is important for understanding NFQ evaluation criteria.}
By analyzing the pointwise method results with and without reference, we find Pointwise$^{R\neq \emptyset}$ always performs better. In some cases, it can even exceed the performance of pairwise methods.
This indicates that providing the reference answer list helps LLMs understand NFQ evaluation criteria so that Pointwise$^{R\neq \emptyset}$ can assign a more reliable score than Pointwise$^{R=\emptyset}$. 
This further illustrates that providing $\mathcal{R}$ in evaluating NFQA can lead to significant gains.

\textbf{LINKAGE is applicable in various of situations.}
Table \ref{1GT} and Table \ref{0GT} show that \baby-1GT and \baby-0GT both perform the best among all LLM evaluation methods. This illustrates that our method is still effective when generalized to other evaluation scenarios, i.e., when there is only one ground truth or no ground truth.

\subsection{Study on the Reference List Composition}
\label{composition of R}
We conduct experiments on different reference distributions to analyze their impact. As shown in Table \ref{composition effect}, varying length and distribution of $\mathcal{R}$ affects the performance of \baby. To ensure the fairness of the experiment, candidate answers are the same for each setting.

The impact of length depends on the quality of the dataset. ANTIQUE is collected from web data and contains more noise, so increasing the number of references can help LLMs better build evaluation criteria. The conclusion on TREC-DL-NF is the opposite. For quality assurance datasets, increasing the number of references, however, exacerbates the burden of understanding long texts, thereby impairing evaluation performance. For the grade distribution of reference answers, uniform sampling always brings the best results, as it allows LLMs to understand all grades of answers while avoiding introducing grade preference bias.



\subsection{Study on the Reference List Randomness}
Our experiments involve random sampling of the ground truth set, so we evaluate the results under 3 randomizations to analyze the impact of randomness on performance.
Results on TREC-DL-NF are in Table~\ref{tab:random trec} and results on ANTIQUE can be found in Appendix \ref{appendix: random}. We can observe that for all LLMs on all datasets, the standard deviations of the experiments are always small. This indicates that the randomness of the selection of reference answers has little impact on the evaluation results, which proves that the improvement brought by our method is significant.

\begin{figure}[t]
    \centering
    \includegraphics[width=\linewidth]{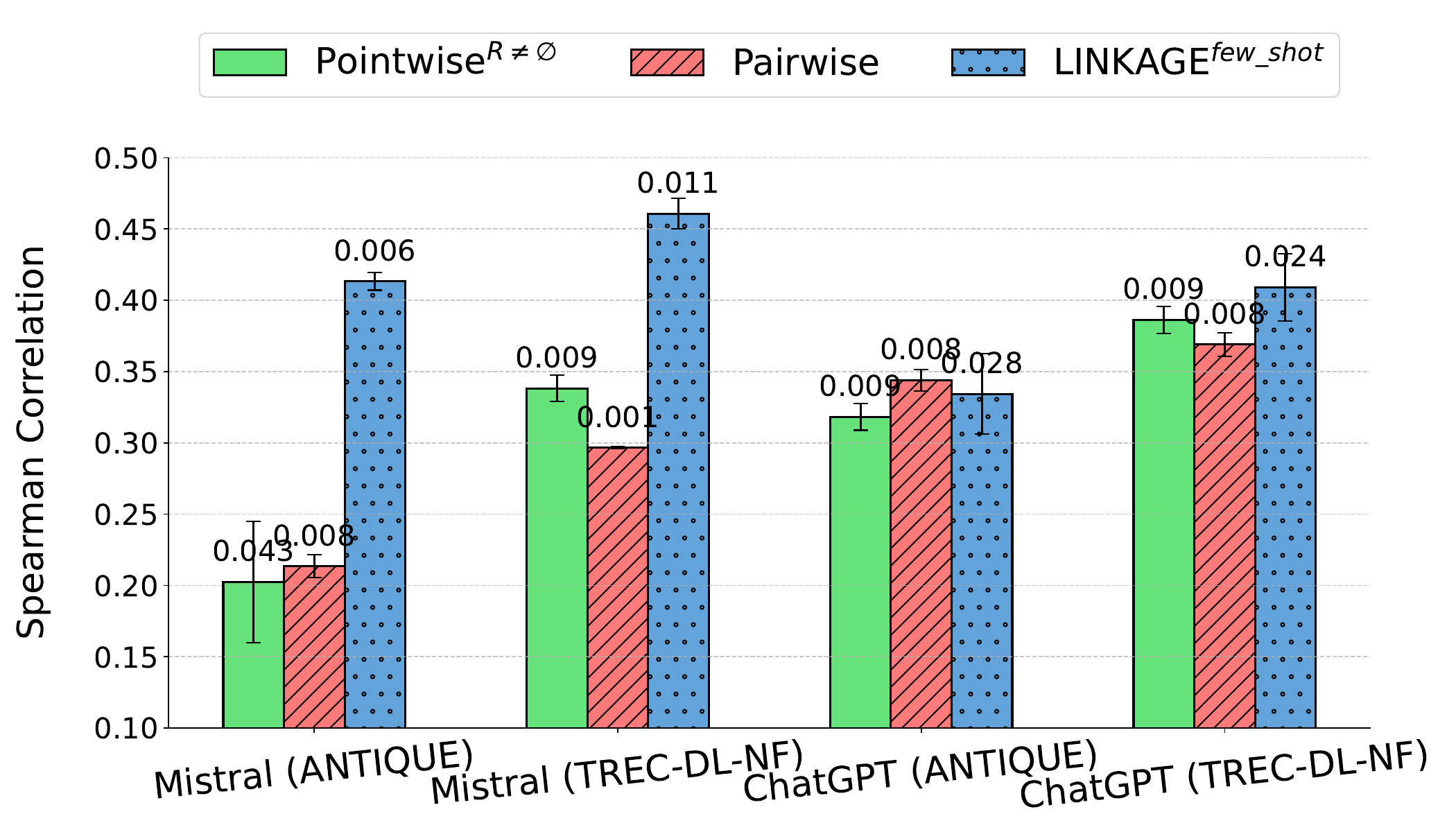}
    \caption{Comparison of Spearman Correlation for Mistral and ChatGPT on ANTIQUE and TREC-DL-NF. The error bars denote the standard deviation, illustrating the variability in the results.}
    \label{fig:random}
\end{figure}

\section{Case Study}
\label{case_study}

\begin{figure*}[t]
    \centering
    \includegraphics[width=\linewidth]{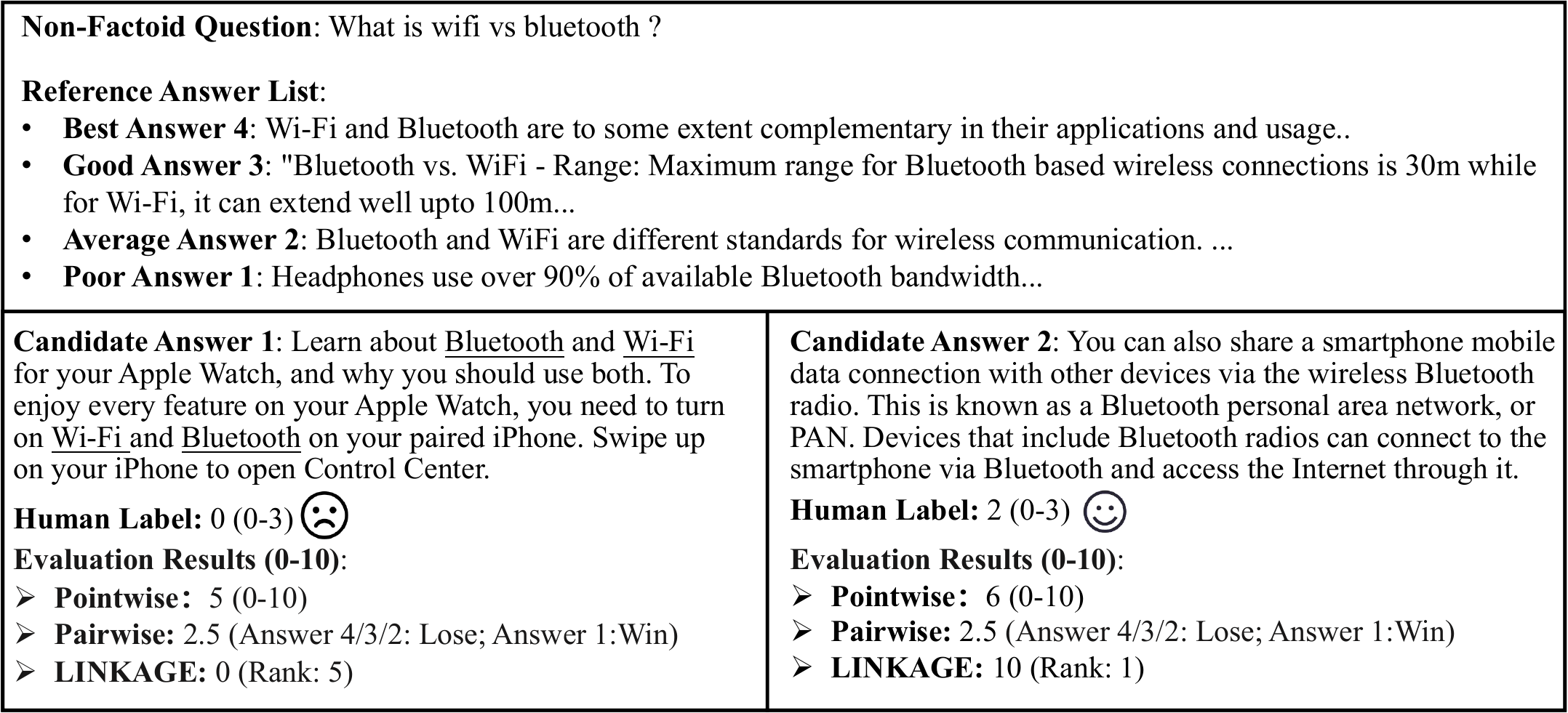}
    \caption{An example of our \baby\, compared with pointwise and pairwise approaches. We standardized the score range of all methods to $[0, 10]$ for easy comparison and understanding.}
    \label{fig:case}
\end{figure*}
We conduct case studies to qualitatively compare the results of different methods.
As shown in the Figure~\ref{fig:case}, because candidate answer 1 contains many matching keywords, even though it does not effectively answer the question, pointwise method and pairwise method both assign it a high score. As a result, the two candidate answers cannot be effectively distinguished. In contrast, our LINKAGE can better distinguish the fine-grained quality differences between candidate answers and obtain results that are more consistent with humans.


\section{Conclusion}

In this paper, we propose a listwise NFQA evaluation approach (\baby), which leverages LLMs to assess a candidate answer by its rank in a list of sorted reference answers.  Our approach is capable of considering reference answers of various quality simultaneously. Therefore, it can enable LLMs to establish a better evaluation system and yield more accurate assessments. Extensive experiments on three datasets, i.e., ANTIQUE, TREC-DL-NF, and WebGLM, demonstrate the effectiveness of our method, whether it is in situations with multi-grade ground truth answers, single-grade ground truth answers, or no ground truth. Hoping this more accurate evaluation method can promote future research on NFQA.
\section*{Limitations}
There are two primary limitations:
\begin{enumerate*}[label=(\roman*)]

    \item Our method demands multiple grading labels when constructing the reference answer list. When grading labels are missing, utilizing LLMs to generate reference answers increases the cost of inference. How to reduce the computational cost requires further research in the future.
    
    \item Compared with the pointwise and pairwise methods, the listwise method considers the relationship between all documents, so it requires the scoring model to have a good long-text understanding ability.
\end{enumerate*}
\section*{Acknowledgement}
This work was funded by the National Natural Science Foundation of China (NSFC) under Grants No. 62302486, the Innovation Project of ICT CAS under Grants No. E361140, the CAS Special Research Assistant Funding Project, the Lenovo-CAS Joint Lab Youth Scientist Project, the project under Grants No. JCKY2022130C039, and the Strategic Priority Research Program of the CAS under Grants No. XDB0680102.
\normalem
\bibliography{reference}
\appendix
\onecolumn
\section{Instruction Details}
\label{sec:appendix}
\subsection{Instruction for Evaluation}
\label{a.1}
The evaluation prompts are adopted in \baby\ and LLM baselines (detailedly introduced in Sec\ref{methods for comparison}).
These prompts are fed to LLMs, allowing them to generate scores, preferences or rankings. 
\begin{figure*}[ht]
    \centering
    \includegraphics[width=\linewidth]{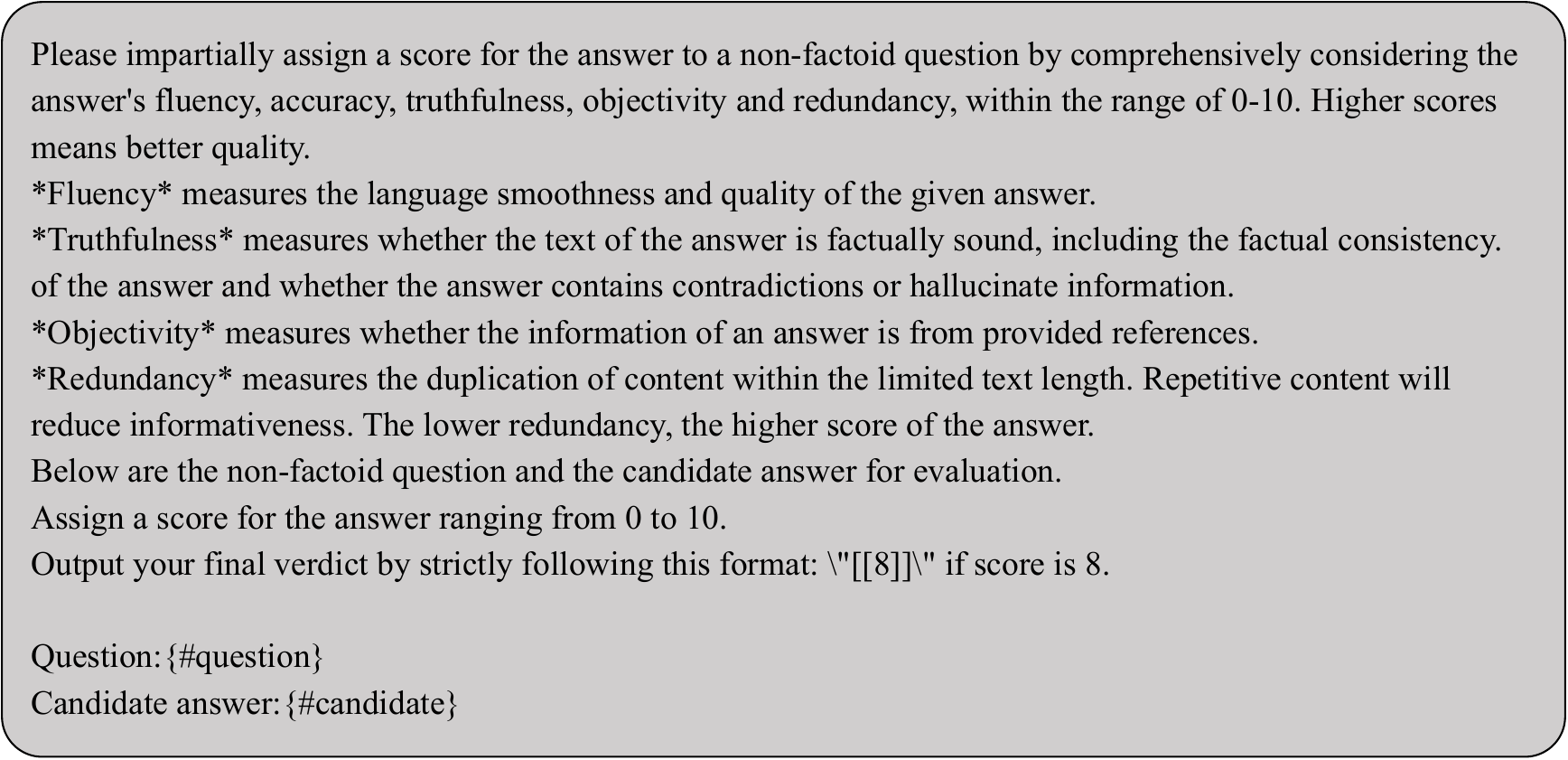}
    \vspace{-3mm}
    \caption{Instruction for pointwise scoring without references.}
\vspace{-1mm}
    \label{fig:point_no_ref}
\end{figure*}

\begin{figure*}[ht]
    \centering
    \includegraphics[width=\linewidth]{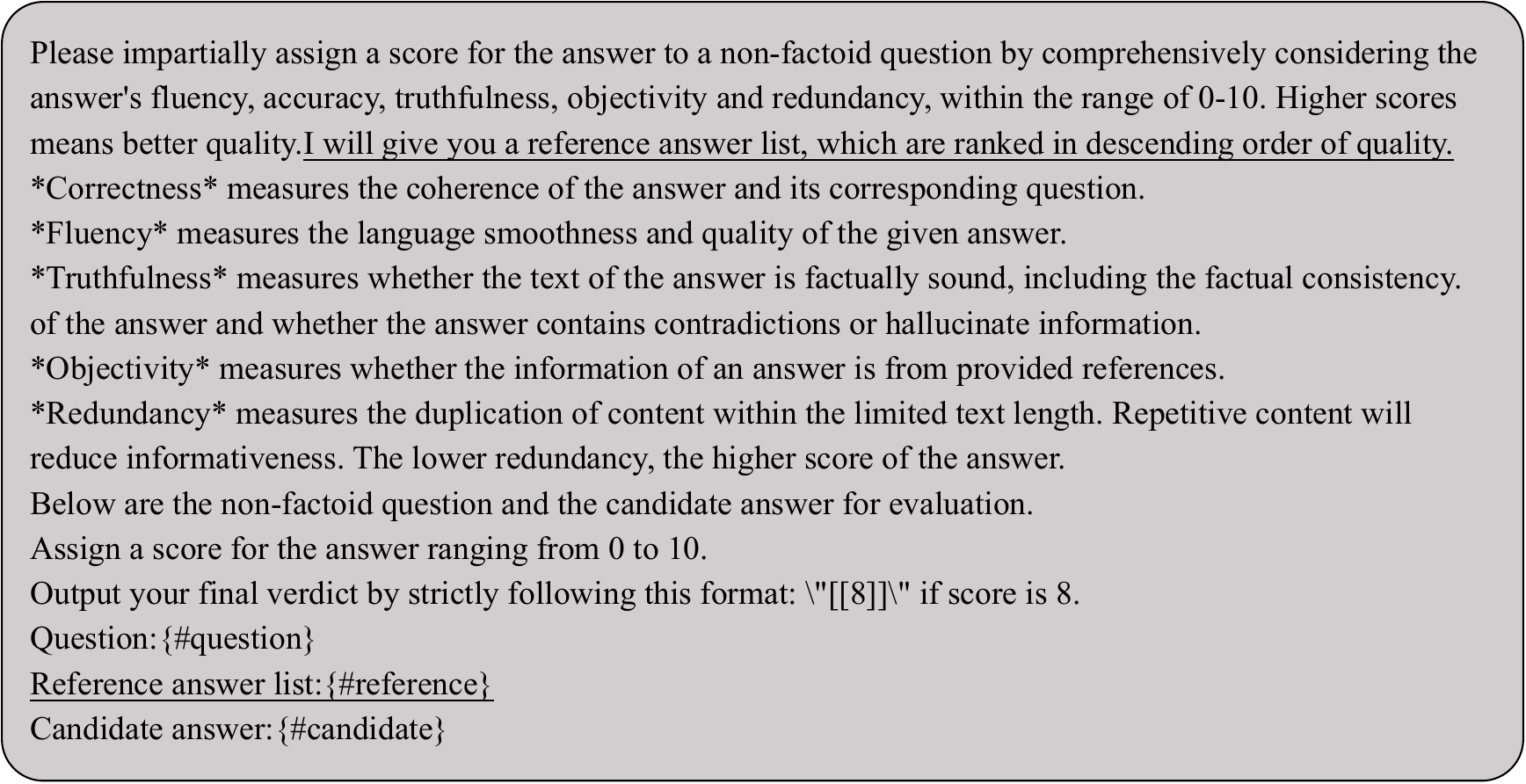}
    \vspace{-3mm}
    \caption{Instruction for pointwise scoring with references.}
\vspace{-1mm}
    \label{fig:point_with_ref}
\end{figure*}

\begin{figure*}[ht]
    \centering
    \includegraphics[width=\linewidth]{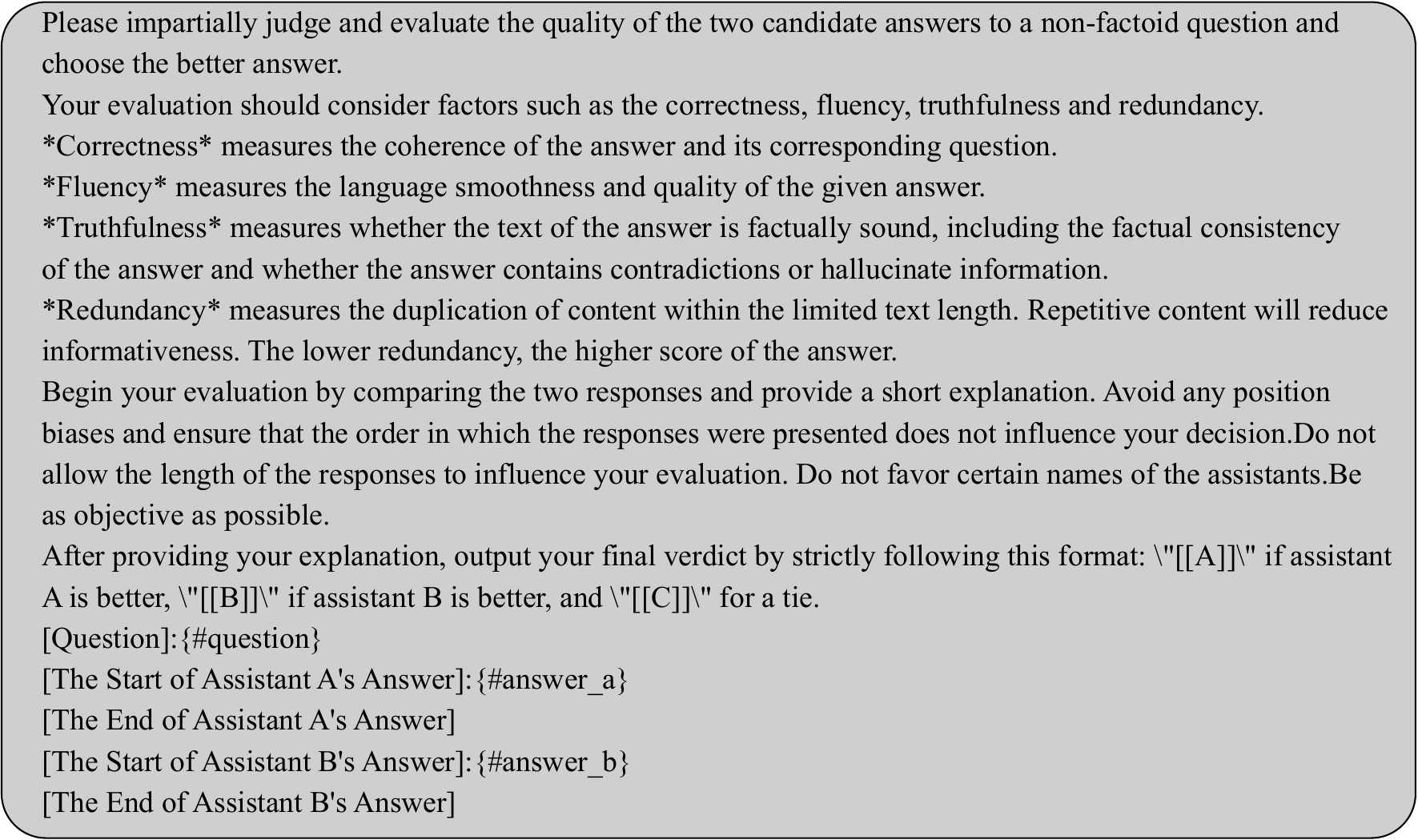}
    \vspace{-3mm}
    \caption{Instruction for pairwise comparison.}
\vspace{-1mm}
    \label{fig:pairwise}
\end{figure*}

\begin{figure*}[ht]
    \centering
    \includegraphics[width=\linewidth]{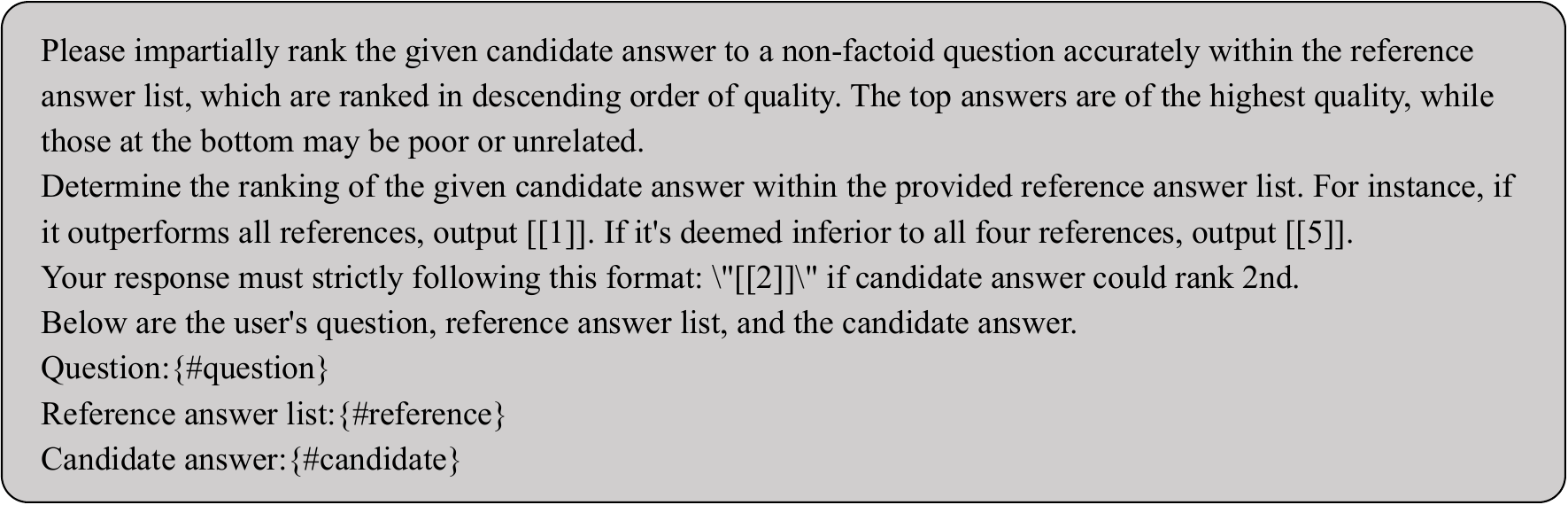}
    \vspace{-3mm}
    \caption{Instruction for our proposed \baby.}
\vspace{-1mm}
    \label{fig:our}
\end{figure*}
\clearpage

\subsection{Instruction for Generating Reference List}
\label{a.2}
\begin{figure*}[ht]
    \centering
    \includegraphics[width=\linewidth]{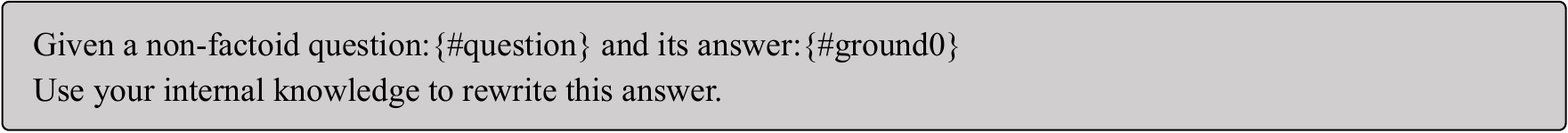}
    \vspace{-3mm}
    \caption{Instruction for generating the highest standard reference answer.}
\vspace{-1mm}
    \label{fig:generate r4}
\end{figure*}

\begin{figure*}[ht]
    \centering
    \includegraphics[width=\linewidth]{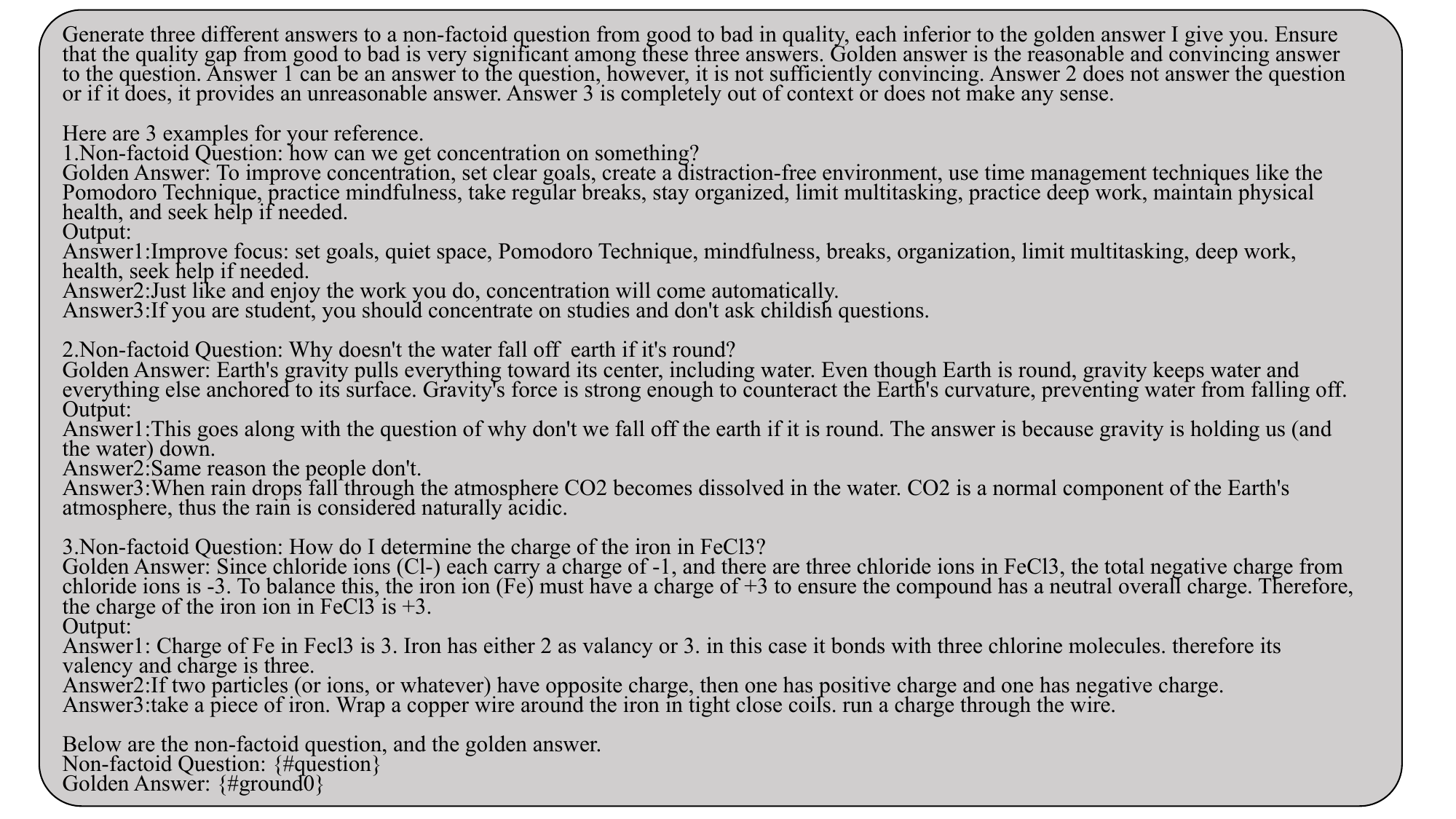}
    \vspace{-3mm}
    \caption{Instruction for generating other reference answers in $\mathcal{R}$ sorted by quality descendingly.}
\vspace{-1mm}
    \label{fig:generate other R}
\end{figure*}
\section{Experiments on Randomness of Reference List}
\label{appendix: random}
Three independent experiments on randomly selecting $\mathcal{R}$ on ANTIQUE using ChatGPT and Mistral are shown in Table \ref{tab:random antique}.


\begin{table*}[ht]
\setlength\tabcolsep{2pt} 
\renewcommand{\arraystretch}{1.2} 
\setlength{\abovecaptionskip}{0pt}
\caption{Results of randomly select $\mathcal{R}$ in three dependent experiments on ANTIQUE.}
\label{tab:random antique}
\centering
\begin{threeparttable}
  \begin{tabular}{lcccccc}
    \toprule
    Model & Method & Spearman 1 & Spearman 2 & Spearman 3 & Average & Std \\
    \midrule
    \multirow{3}{*}{Mistral} 
    & Pointwise$^{R\neq \emptyset}$ & 0.2516 & 0.1781 & 0.1778 & 0.2025 & 0.0425\\
    & Pairwise &0.2210 & 0.2059 & 0.2134 & 0.2134& 0.0082\\
    & \baby$^{few\_shot}$ & 0.4200 & 0.4078 & 0.4122 & 0.4133 & 0.0062\\
    \midrule
    \multirow{3}{*}{ChatGPT} 
    & Pointwise$^{R\neq \emptyset}$ & 0.3118 & 0.3250 & 0.3180 & 0.3182 & 0.0094\\
    & Pairwise &0.3495 & 0.3387 & 0.3402 & 0.3428 & 0.0076\\
    & \baby$^{few\_shot}$ & 0.3543 & 0.3143 & 0.3339 & 0.3340 & 0.0283\\
    \bottomrule
  \end{tabular}
\end{threeparttable}

\end{table*}
\clearpage

\section{Case Study}
\label{appendix:case}
\begin{figure*}[ht]
    \centering
    \includegraphics[width=\linewidth]{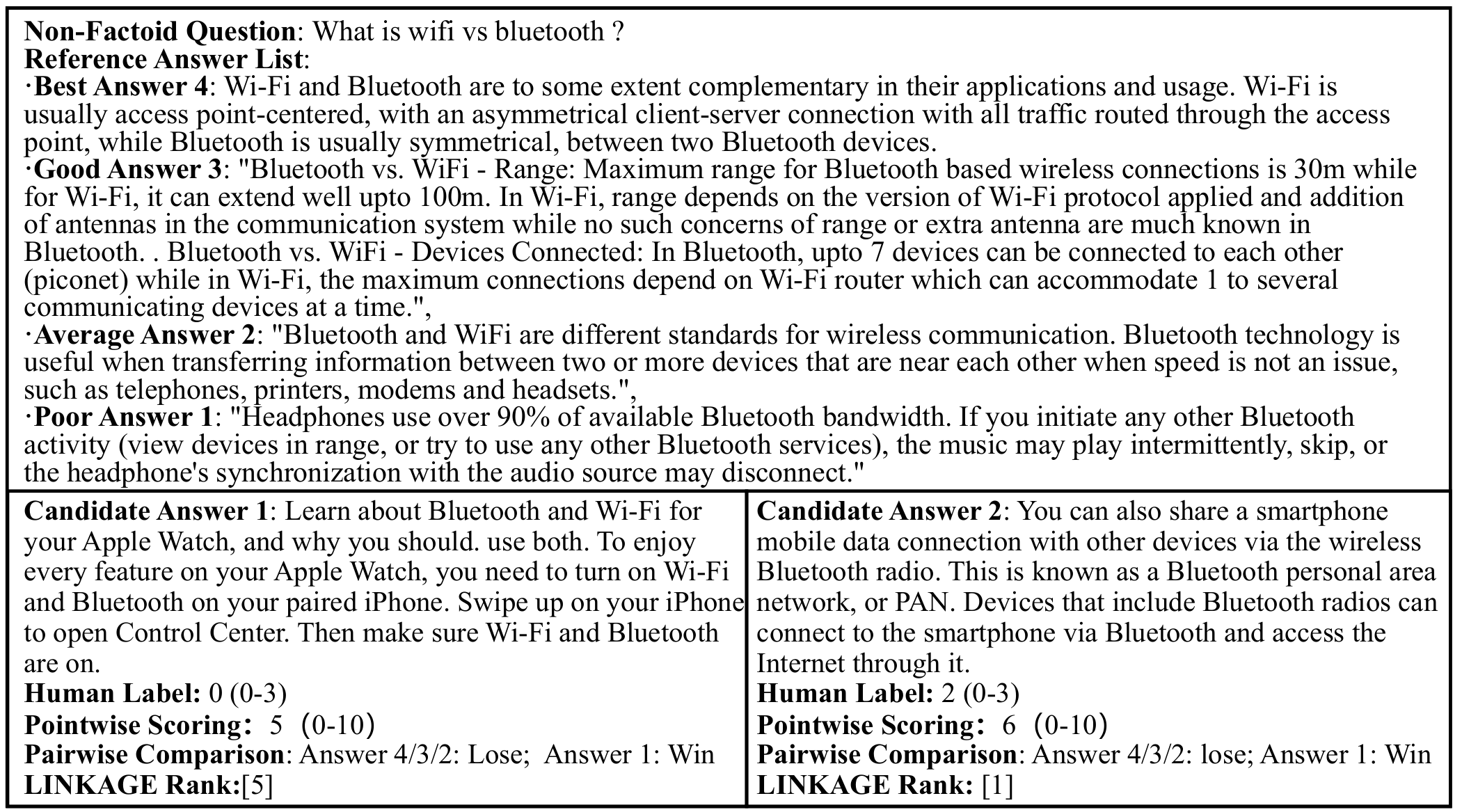}
    \vspace{-3mm}
    \caption{An example of our \baby\, compared with Pointwise and Pairwise approach.}
\vspace{-1mm}
    \label{fig:detail_case}
\end{figure*}

The details of case in Section~\ref{case_study} of the main paper is in Figure \ref{fig:detail_case}.

\clearpage
\section{Human Annotation}
\begin{figure*}[ht]
    \centering
    \includegraphics[width=\linewidth]{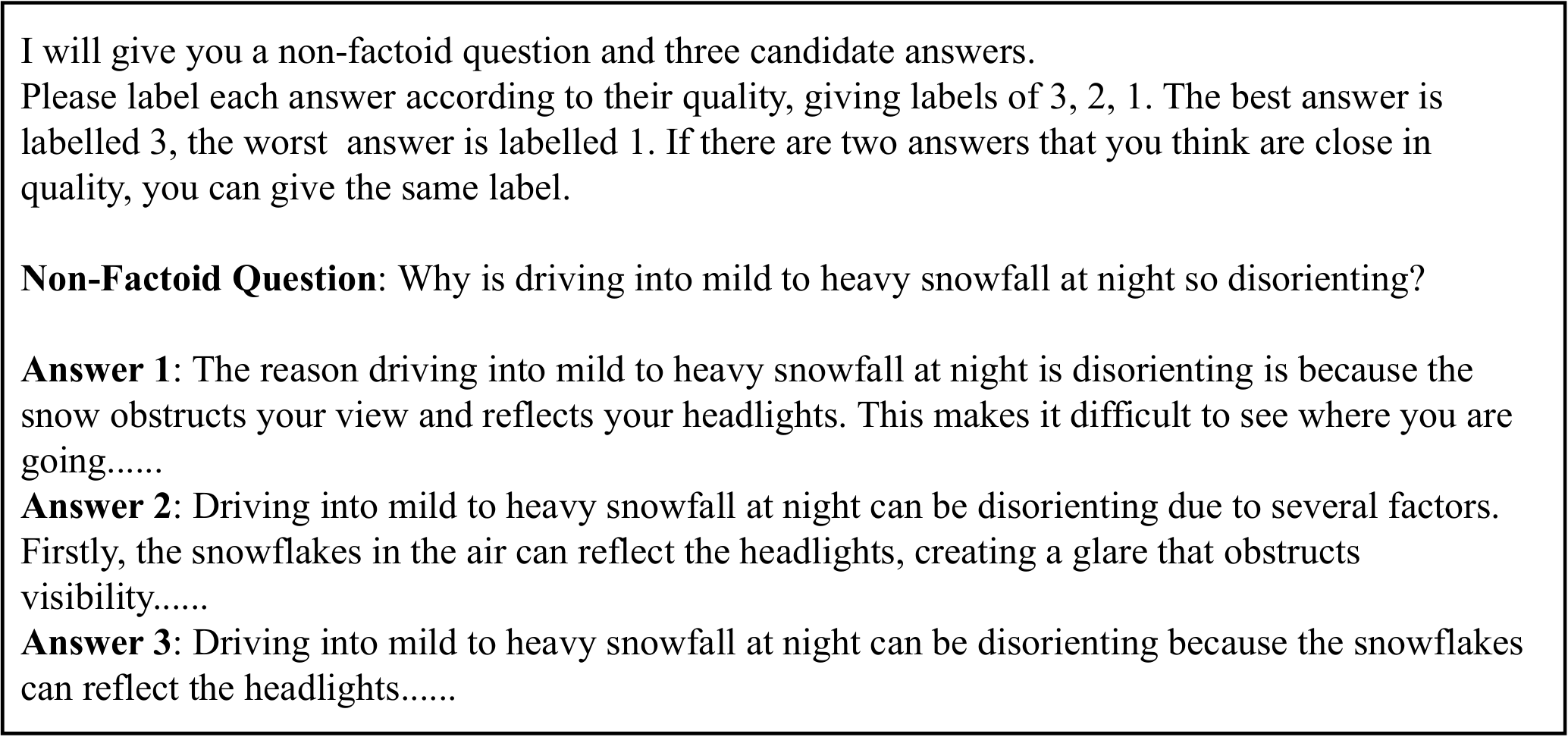}
    \vspace{-3mm}
    \caption{Instructions for labeling WEBGLM for human annotators.}
\vspace{-1mm}
    \label{fig:human}
\end{figure*}

We recruit one domain expert who has earned at least a bachelor's degree in Computer Science to annotate WEBGLM candidate answer's quality label. The instruction is shown in Figure \ref{fig:human}.
\label{appendix:human}

\end{document}